\documentclass{article} % For LaTeX2e
\usepackage[preprint]{colm2025_conference}

\usepackage{microtype}
\usepackage{hyperref}
\usepackage{url}
\usepackage{booktabs}

\usepackage{lineno}

\usepackage{amsfonts}       % blackboard math symbols
\usepackage{nicefrac}       % compact symbols for 1/2, etc.
\usepackage{microtype}      % microtypography
\usepackage{xcolor}         % colors
\usepackage{amsmath}

\usepackage{pifont}
\usepackage{amsmath}
\usepackage{algorithm}
\usepackage{algpseudocode}
\usepackage{wrapfig}
\usepackage{graphicx}
\usepackage{booktabs}
\usepackage{wrapfig}
\usepackage{subcaption}
\usepackage{multirow}
\usepackage{makecell}

\definecolor{darkblue}{rgb}{0, 0, 0.5}
\hypersetup{colorlinks=true, citecolor=darkblue, linkcolor=darkblue, urlcolor=darkblue}

\title{DistZO2: High-Throughput and Memory-Efficient Zeroth-Order Fine-tuning LLMs with Distributed Parallel Computing}

\author{Liangyu Wang \\
King Abdullah University of Science and Technology (KAUST) \\
Saudi Arabia \\
\texttt{\{liangyu.wang\}@kaust.edu.sa} \\
\And
Huanyi Xie \\
King Abdullah University of Science and Technology (KAUST)\\
Saudi Arabia \\
\texttt{\{huanyi.xie\}@kaust.edu.sa} \\
\AND
Di Wang \\
King Abdullah University of Science and Technology (KAUST)\\
Saudi Arabia \\
\texttt{\{di.wang\}@kaust.edu.sa} \\
}

\begin{document}

\ifcolmsubmission
\linenumbers
\fi

\maketitle

\begin{abstract}
Fine-tuning large language models (LLMs) remains resource-intensive due to their sheer scale. While zeroth-order (ZO) optimization provides a memory-efficient alternative by eliminating backward passes, its application to multi-hundred-billion-parameter models is constrained by GPU memory and compute throughput. The ZO2 framework addresses the memory bottleneck by offloading model parameters to CPU memory and overlapping transformer block transfer with dual forward computation on a single GPU. However, ZO2 remains limited by its single-device execution and achieves modest throughput.
In this work, we present DistZO2, a high-throughput, memory-efficient framework for distributed zeroth-order fine-tuning of LLMs. DistZO2 introduces three parallel strategies: (1) Perturbation Parallelism (PertP), which parallelizes the two perturbed forward passes across devices; (2) Distributed Data Parallelism (DDP), adapted to the scalar-gradient nature of ZO training; and (3) a unified 2D Parallelism design that combines PertP and DDP. To further mitigate communication bottlenecks introduced by parameter offloading, we propose a hardware-aware communication strategy that slices parameter blocks and redistributes them across GPUs via high-speed interconnects such as NVLink.
DistZO2 scales zeroth-order fine-tuning to modern multi-GPU systems, preserving ZO2’s memory efficiency while substantially improving training throughput.
In our experiments on OPT-175B, DistZO2 achieves a $\mathbf{3\times}$ speedup over ZO2 with distributed computing. DistZO2's code has been open-sourced in \href{https://github.com/liangyuwang/zo2}{\textit{https://github.com/liangyuwang/zo2}}.
\end{abstract}

\section{Introduction}

Large Language Models (LLMs) have demonstrated remarkable capabilities across a wide range of tasks, but fine-tuning them remains prohibitively expensive. For instance, models such as GPT-175B \citep{brown2020languagemodelsfewshotlearners}, OPT-175B \citep{zhang2022opt}, LLaMA 405B \citep{dubey2024llama3}, and DeepSeek 671B \citep{deepseekai2025deepseekv3technicalreport} require substantial GPU memory and computational resources, making full-parameter fine-tuning challenging in practical settings.

Recent work has shown that zeroth-order (ZO) optimization offers a promising alternative for fine-tuning LLMs without requiring backward passes \citep{malladi2023fine}. By estimating gradients through forward-only computations, ZO methods eliminate the need to store intermediate activations, drastically reducing memory consumption. However, naively applying ZO fine-tuning to extremely large models remains limited by GPU memory and compute throughput.

To address the memory bottleneck, the ZO2 framework \citep{wang2025zo2scalablezerothorderfinetuning} enables zeroth-order fine-tuning of multi-hundred-billion-parameter LLMs (e.g., OPT-175B) using a single GPU, by offloading most parameters to CPU memory. ZO2 introduces a fine-grained task scheduler to effectively overlap CPU–GPU communication with forward computation. Nevertheless, ZO2 is fundamentally constrained by single-device compute throughput; for example, it reports a maximum of 37 tokens/second when fine-tuning OPT-175B on a single A100 GPU.

To improve throughput, distributed computing can be leveraged to parallelize forward computation across multiple devices. In particular, ZO methods perform two independent and symmetric forward passes with perturbed parameters in the same direction, enabling natural parallel execution across devices. This contrasts with first-order (FO) methods \citep{ruder2017overviewgradientdescentoptimization,loshchilov2017decoupled}, where the forward and backward passes are dependent and must be executed sequentially. We propose a new strategy called \textit{Perturbation Parallelism (PertP)}, which distributes these two forward passes across two devices. While conceptually straightforward, PertP introduces nontrivial scheduling challenges, especially in managing asynchronous execution, random number synchronization, and overlapping computation with communication.

Another avenue for improving throughput is \textit{Distributed Data Parallelism (DDP)} \citep{li2020pytorchdistributedexperiencesaccelerating}, which scales training by replicating models across multiple GPUs and synchronizing gradients via \texttt{all\_reduce}. However, DDP is traditionally tailored to first-order optimizers (FO), where gradient synchronization happens after a backward pass. In ZO-based fine-tuning, there are no backward passes, and the model produces only scalar projected gradients. Since DDP was originally designed for first-order optimizers, applying it to zeroth-order training requires some adaptation, such as synchronizing scalar projected gradients instead of full tensors and ensuring consistent perturbation vectors across devices.

Moreover, naive distributed communication can degrade performance if not carefully optimized. In particular, when applying distributed computing to ZO2, the CPU must transmit the same $M$-sized parameter block to each GPU, putting significant pressure on limited PCIe bandwidth. ZO2's original single-GPU design overlaps dual forward computation with CPU–GPU data movement, but PertP separates these computations across devices, potentially breaking this overlap. The problem is further exacerbated by advances in compute optimizations—such as FlashAttention \citep{dao2023flashattention2fasterattentionbetter} and \texttt{torch.compile} \citep{ansel2024pytorch}—which can accelerate forward passes to the point where communication becomes the bottleneck.

To address these challenges, we propose DistZO2, a distributed zeroth-order fine-tuning framework that achieves high throughput and low memory overhead. DistZO2 introduces three distributed strategies with communication optimization: (1) \textbf{Perturbation Parallelism (PertP)} to parallelize dual-forward computation, (2) \textbf{Distributed Data Parallelism (DDP)} to scale across batches and devices, and (3) \textbf{2D Parallelism}, a fusion of PertP and DDP into a unified framework.

Our contributions are summarized as follows:
\begin{itemize} 
\vspace{-0.1in}
    \item We identify the key limitations of single-device ZO2 training and propose Perturbation Parallelism (PertP) to exploit the dual-forward structure for inter-device compute parallelism. \item We demonstrate how to adapt Distributed Data Parallelism (DDP) for ZO2 training, including scalar gradient synchronization and consistent RNG (random number generator) seeding. 
    \item We propose a communication optimization strategy that reduces PCIe bottlenecks by slicing and aggregating parameter blocks across GPUs with NVLink coordination. 
    \item We introduce a unified 2D parallelism framework that combines PertP and DDP, enabling scalable and high-throughput zeroth-order fine-tuning on modern multi-GPU systems. 
    \item Through extensive experiments, we show that PertP and DDP respectively address ZO2's compute and data bottlenecks, while our 2D parallelism and communication optimization further amplify throughput—enabling \textbf{3}$\times$ speedup over ZO2 and scalable fine-tuning up to OPT-175B.
\end{itemize}

\section{Related Work}

Zeroth-order (ZO) optimization offers a memory-efficient alternative to first-order methods by eliminating backward passes. MeZO~\citep{malladi2023fine} and ZO2~\citep{wang2025zo2scalablezerothorderfinetuning} demonstrate the feasibility of fine-tuning multi-billion-parameter LLMs using forward-only computation and CPU offloading. However, both are constrained by single-device throughput. Distributed Data Parallelism (DDP)~\citep{li2020pytorchdistributedexperiencesaccelerating} is widely used in first-order training~\citep{ruder2017overviewgradientdescentoptimization, loshchilov2017decoupled}, but require careful adaptation for ZO methods due to the use of scalar projected gradients and the need for consistent perturbations. DistZO2 addresses these challenges by first introducing Perturbation Parallelism (PertP), and then adapting DDP to support ZO training. Furthermore, it proposes a unified 2D parallelism framework that combines PertP with DDP, and alleviates communication bottlenecks through hardware-aware parameter slicing. This design enables high-throughput, memory-efficient fine-tuning of extremely large LLMs. The full related work section can be seen in Appendix~\ref{sec:related-work}.

\section{Preliminaries}

Zeroth-order optimization (ZO) is a gradient-free optimization technique that estimates the directional derivative of the loss function using only forward computations. Given a loss function $\mathcal{L}(\theta)$ and a random direction vector $z \sim \mathcal{N}(0, I)$, the projected gradient, which is a \underline{scalar}, is approximated as:
\begin{equation}
    \label{eq:projected-grad}
    g = \frac{\mathcal{L}(\theta + \epsilon z) - \mathcal{L}(\theta - \epsilon z)}{2\epsilon},
\end{equation}
where $\epsilon$ is a small positive scalar controlling the perturbation magnitude. This estimator is known as the central difference estimator. Then, the model parameters can be updated by:
\begin{equation}
    \label{eq:zo-update}
    \theta = \theta - \eta g z,
\end{equation}
where $\eta$ is the learning rate.
Since ZO does not require backpropagation or storage of intermediate activations, it significantly reduces the memory footprint compared to first-order methods.

Recently, ZO has been applied to the fine-tuning of large language models (LLMs), where the prohibitive memory cost of backpropagation presents a major bottleneck. The MeZO algorithm \citep{malladi2023fine} (Algorithm~\ref{alg:mezo}) demonstrated that ZO can achieve competitive fine-tuning quality using only forward passes. However, applying ZO naively to extremely large models is still infeasible on commodity GPUs due to the size of model parameters.

To address this, the ZO2 framework \citep{wang2025zo2scalablezerothorderfinetuning} (Figure~\ref{fig:zo2}, Algorithm~\ref{alg:zo2}) proposes a memory-efficient zeroth-order fine-tuning system that enables the use of ZO for decoder-only models as large as OPT-175B on a single GPU. ZO2 introduces a block-wise CPU–GPU offloading strategy, where model parameters are stored in host (CPU) memory, and only a single transformer block is loaded to GPU memory at a time for computation. This is enabled by a dual forward computation pattern (positive and negative perturbations) and a simple task scheduler (Algorithm~\ref{alg:zo2-scheduler}) that overlaps block upload, computation, and block offload across CUDA streams. 

However, ZO2 remains limited by its single-GPU design, which constrains training throughput. This motivates our exploration of distributed zeroth-order fine-tuning strategies to improve scalability and throughput further.

\section{Perturbation Parallelism (PertP)} \label{sec:pertp}

\begin{figure}[htbp]
    \centering
    \includegraphics[width=\linewidth]{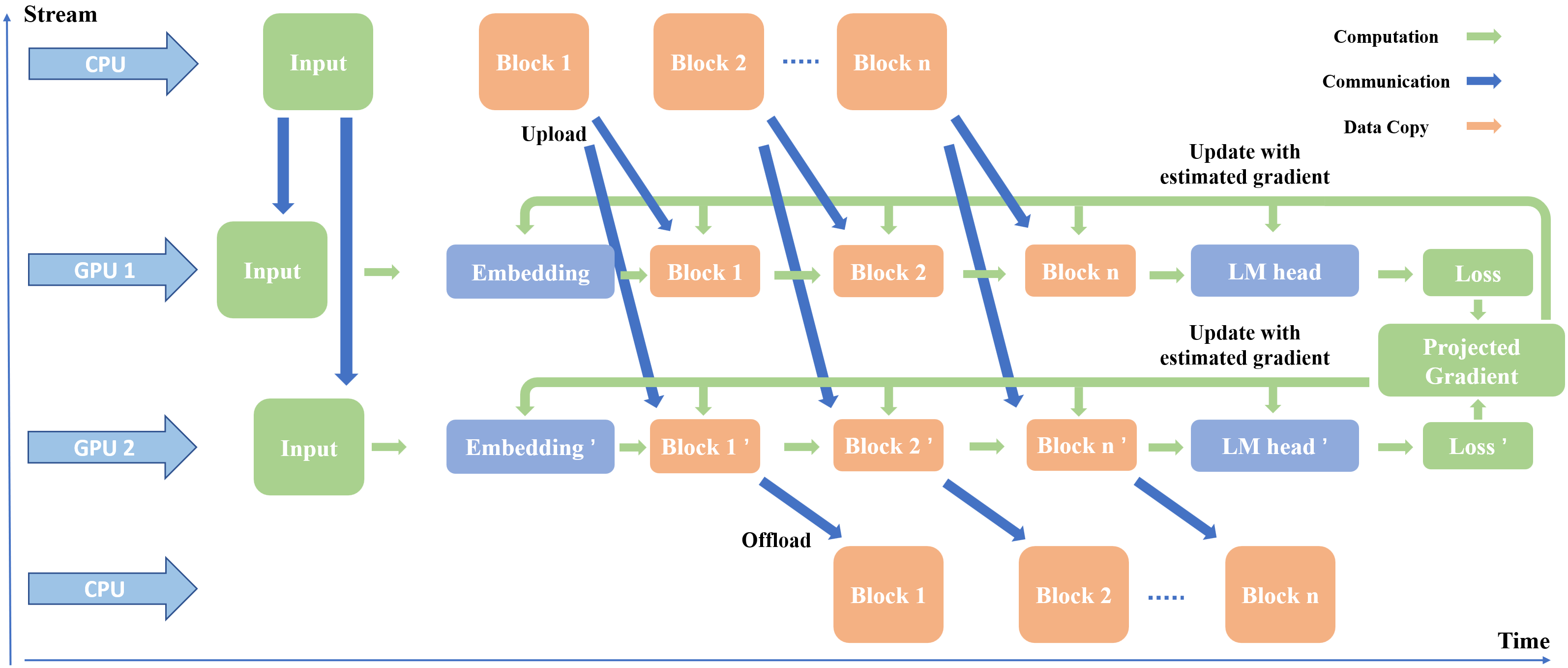}
    \caption{Perturbation Parallelism Workflow.}
    \label{fig:pertp}
\end{figure}

\subsection{Motivation}

The ZO2 framework achieves remarkable memory savings by offloading model blocks to the CPU and avoiding backward passes via ZO. However, its training throughput is limited by the need to perform two complete forward computations per iteration—one under positively perturbed parameters and the other under negatively perturbed ones. In the original ZO2 implementation, these dual forward passes are executed sequentially on a single GPU.

Importantly, this sequential execution is not due to algorithmic dependencies but rather stems from resource constraints: for extremely large models, each forward pass through the transformer layers saturates the GPU's compute and memory bandwidth. As a result, even if the two passes are logically independent, they must be performed one after the other on a single GPU to avoid resource contention.

This observation motivates us to ask: Can we leverage the independence of the two forward passes to improve throughput, if we have access to two GPUs? Unlike first-order training, where the forward and backward passes are tightly coupled and sequential by nature, zeroth-order methods offer a natural opportunity for parallelism. We introduce Perturbation Parallelism (PertP) to exploit this opportunity.

\subsection{Workflow}

PertP is a distributed training strategy designed specifically for zeroth-order optimization. It leverages the structural independence between the two forward passes required for estimating zeroth-order gradients: one using positively perturbed parameters, and the other using negatively perturbed parameters. Rather than executing both computations sequentially on a single GPU, PertP assigns them to two separate GPUs, allowing the computations to run concurrently.

Let $W \in \mathbb{R}^{p \times d}$ denote the parameter matrix of a transformer block. In the ZO2 framework, the master copy of this block is stored on the CPU, denoted as $W_H$, and is dynamically transferred to GPU memory when needed. Let $W_{D_k}$ be the device-side copy on GPU $k \in \{1,2\}$. Each GPU performs a forward pass using perturbed versions of $W_{D_k}$ to compute scalar losses for gradient estimation.

PertP retains ZO2’s core features, such as CPU offloading and communication-computation overlap. Each GPU independently uploads transformer blocks from the CPU, performs its assigned forward pass, and offloads the block afterward. Once both losses are computed, a lightweight communication step merges the results to form the final projected gradient.

Figure~\ref{fig:pertp} illustrates the system architecture and execution flow. Each training iteration in PertP consists of the following steps:

\textbf{Step 1: Input Preparation and Shared RNG Initialization.}
At the start of each iteration, a mini-batch is loaded into CPU memory and broadcast to both GPUs. A shared random seed $s$ is generated and broadcast to ensure both devices sample the same Gaussian perturbation vector $z \in \mathbb{R}^{p \times d}$.

\textbf{Step 2: Distributed Dual Forward Execution.}
Each GPU independently performs a forward pass on its local copy of the parameters. Specifically:
(1) GPU 1 computes the forward pass with positively perturbed parameters: $W_{D_1}^+ = W_H + \epsilon z$.
(2) GPU 2 computes the forward pass with negatively perturbed parameters: $W_{D_2}^- = W_H - \epsilon z$.
Each GPU computes its own scalar loss, denoted as $L_1^+$ and $L_2^-$, respectively.

\textbf{Step 3: Loss Synchronization and Projected Gradient Computation.}
Once both losses are available, they are exchanged between devices using \texttt{all\_gather} or \texttt{send/recv}. Then, each device computes the projected gradient using:
$g = \frac{L_1^+ - L_2^-}{2\epsilon}$.
This scalar gradient is identical across devices and can be reused for all updated parameters.

\textbf{Step 4: In-Place Parameter Update.}
Each GPU applies the projected gradient $g$ to its local parameters using the shared direction vector $z$:
$W_{D_k} \leftarrow W_{D_k} - \eta \cdot g \cdot z$.
Since both $g$ and $z$ are synchronized, the updates are consistent across devices without further communication.

\textbf{Step 5: Overlapping Computation and Communication.}
To maximize hardware utilization, each GPU maintains asynchronous CUDA streams for uploading, computing, and offloading transformer blocks. While a block is being processed, the next block is pre-loaded, and the previous one is asynchronously offloaded. These operations are executed independently per GPU.

\section{Applying Distributed Data Parallelism (DDP) to ZO2} \label{sec:ddp}

\begin{figure}[htbp]
    \centering
    \includegraphics[width=\linewidth]{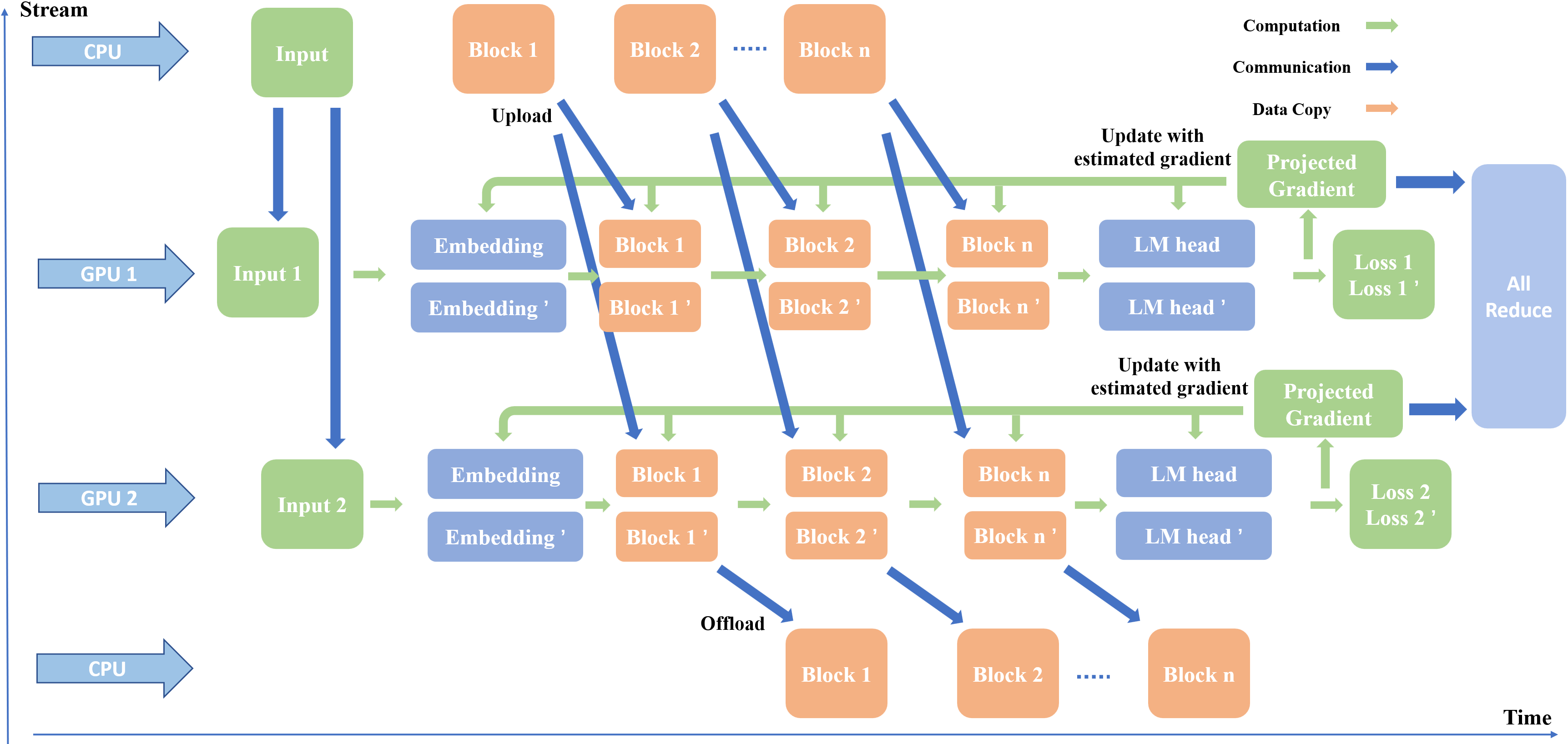}
    \caption{Applying Distributed Data Parallelism to ZO2. Each GPU performs dual forward passes on its own batch, computes a scalar projected gradient, and synchronizes it via \texttt{all\_reduce}.}
    \label{fig:ddp}
    \vspace{-0.15in}
\end{figure}

\subsection{Motivation and Workflow}

The ZO2 framework enables large-scale zeroth-order (ZO) fine-tuning using a single GPU by offloading model parameters to CPU memory and avoiding backward passes. However, the throughput remains limited by sequential data processing and per-device compute capacity. To further scale training throughput, we introduce a distributed version of ZO2 using \textit{Distributed Data Parallelism (DDP)}.

Applying DDP to ZO2 differs from its use in first-order (FO) optimization. In FO methods, each GPU computes high-dimensional gradients via backpropagation, and \texttt{all\_reduce} is used to synchronize full gradient tensors. In contrast, ZO-based training produces only a \textit{scalar} projected gradient per iteration. This makes gradient synchronization highly efficient in the ZO setting and allows DDP to scale with minimal communication overhead.

In each iteration, every GPU processes a different mini-batch and computes the ZO2 dual forward passes to estimate a local projected gradient $g_k \in \mathbb{R}^1$. To ensure model consistency across devices, we use an \texttt{all\_reduce} to aggregate scalar gradients: $g_{\text{avg}} = \frac{1}{K} \sum_{k=1}^{K} g_k$.

In addition, all GPUs must use the same perturbation vector $z$ for gradient estimation. We broadcast a shared random seed from the primary rank (e.g., GPU 0) to all other devices before sampling $z$, ensuring consistent perturbation across devices.

\noindent
\textbf{Workflow Summary.}  
The complete DDP-enabled ZO2 workflow is as follows:
(1) The input mini-batch is partitioned across $K$ GPUs.
(2) Each GPU uploads relevant transformer blocks from CPU to GPU memory, then performs dual forward passes under $\theta \pm \epsilon z$.
(3) Each GPU computes a local scalar projected gradient $g_k$.
(4) Gradients $\{g_k\}$ are averaged via \texttt{all\_reduce} to obtain global $g$.
(5) Each GPU applies the same gradient $g$ and direction vector $z$ to update parameters, and offloads updated blocks back to CPU.

\vspace{-0.15in}
\section{Further Optimization}
\vspace{-0.15in}

\subsection{Communication Optimization Strategy} \label{sec:comm}
\vspace{-0.1in}

One of the key strengths of ZO2 lies in its ability to overlap compute-intensive dual forward passes with communication operations such as block uploads and offloads. However, in the PertP setting, the dual forward pass is decomposed and distributed across GPUs, making it more difficult to maintain this natural overlap between computation and communication. Furthermore, in DDP configurations, a single CPU thread is often responsible for transferring parameter blocks to multiple GPUs simultaneously over PCIe. This can lead to transfer contention and bandwidth saturation, further increasing the overall communication latency. To address this challenge, we propose a communication optimization strategy that leverages the hardware topology of modern GPU clusters to redistribute and parallelize data transfer more efficiently.

\begin{wrapfigure}{r}{0.7\textwidth}
\vspace{-0.2in}
  \begin{center}
    \begin{subfigure}{0.34\textwidth}
      \includegraphics[width=\linewidth]{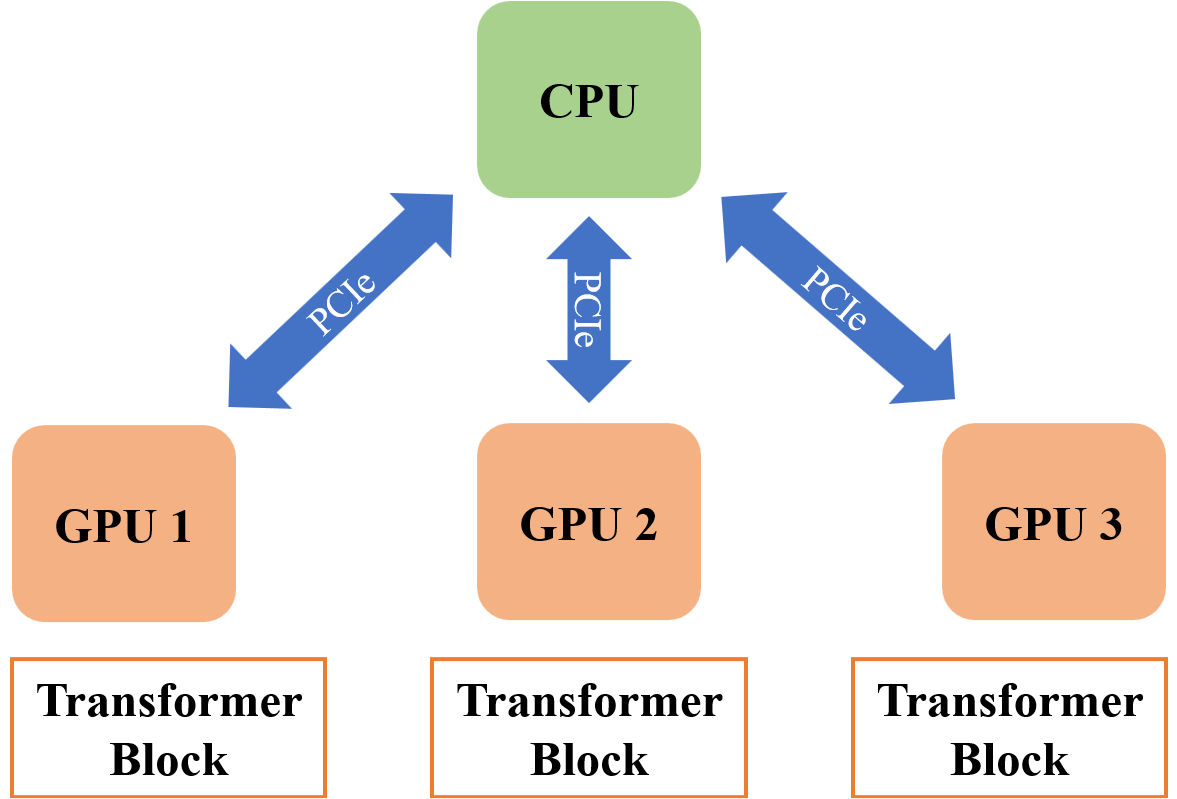}
      \caption{Naive communication. H2D (host to device): each PCIe transfer uploads $M$ parameters. D2H (device to host): each PCIe transfer offloads $M$ parameters.}
      \vspace{0.2in}
    \end{subfigure}
    \hfill
    \begin{subfigure}{0.34\textwidth}
      \includegraphics[width=\linewidth]{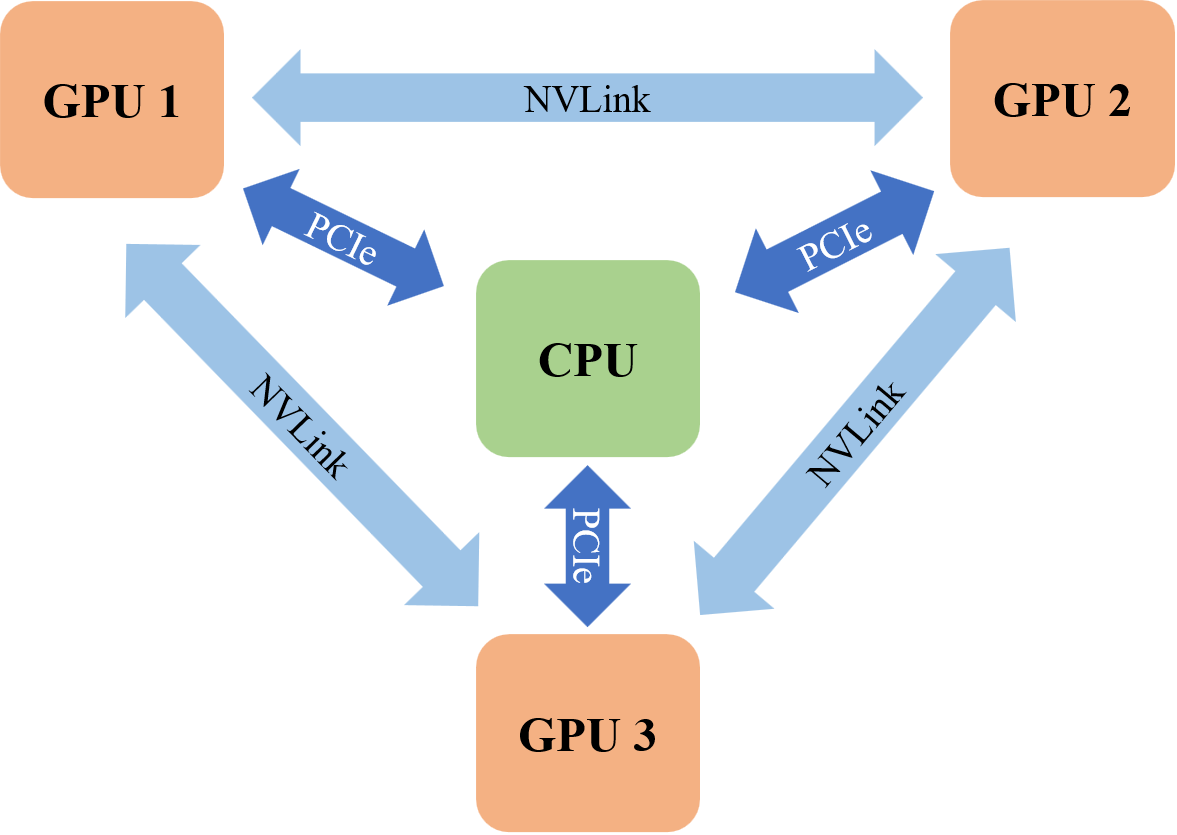}
      \caption{Our solution. H2D: each PCIe or NVLink transfer handles $\frac{M}{n}$ parameter slices, with NVLink transfers initiating only after PCIe transfers are complete. D2H: each PCIe transfer $\frac{M}{n}$ parameters. }
    \end{subfigure}
  \end{center}
  \vspace{-0.1in}
  \caption{Our Communication Optimization Strategy.}
  \label{fig:comm}
\vspace{-0.1in}
\end{wrapfigure}

\textbf{Hardware-Aware Communication Design.}
As illustrated in Figure~\ref{fig:comm}, the CPU is typically connected to GPUs via relatively slow PCIe links, while GPUs within the same node are connected via high-speed NVLink. In naive ZO2 communication (Figure~\ref{fig:comm}a), each GPU independently receives a full-size transformer block of $M$ parameters via PCIe, resulting in redundant data transfers and congested PCIe bandwidth.

\textbf{Optimized Upload Strategy.}
Our strategy (Figure~\ref{fig:comm}b) avoids this inefficiency by adopting the fine-grained tensor slicing and peer-to-peer communication strategies inspired by prior work on distributed tensor communication \citep{shoeybi2020megatronlmtrainingmultibillionparameter, rajbhandari2020zeromemoryoptimizationstraining, jia2018datamodelparallelismdeep, zheng2022alpaautomatinginterintraoperator}: the CPU slices each transformer block into $n$ equal parts, each of size $M/n$, and sends one slice to each of the $n$ GPUs via PCIe. After receiving its assigned slice, each GPU shares the remaining $(n-1)$ slices with its peers through NVLink-based peer-to-peer communication. This design reduces PCIe load per GPU by a factor of $n$, and utilizes the faster NVLink interconnect for intra-node communication. Since PCIe and NVLink communication streams can be executed concurrently, the total upload time per GPU becomes:
$T_{\text{comm}} = \frac{M}{n}\text{ (PCIe)} + \frac{(n-1)M}{n} \text{ (NVLink)}$.

\textbf{Optimized Offload Strategy.}
After projected gradient computation, updated parameters must be transferred back to CPU memory. In naïve ZO2 offloading, each GPU would offload its entire copy of the updated parameter block, resulting again in redundant PCIe traffic. However, due to the deterministic and synchronized nature of zeroth-order updates, all GPUs hold identical updated parameters after the all-reduce operation. Thus, it is sufficient for each GPU to offload only its $M/n$ slice. The CPU reassembles the full $M$-sized block by concatenating the $n$ slices, with no loss of correctness.

\textbf{Thread-Aligned Memory Partitioning.}
To avoid repeated slicing and merging of parameter blocks, we align model initialization with the communication pattern. During model setup, each CPU thread (assigned to control one GPU) is allocated a persistent memory region responsible for its $M/n$ slice. This avoids the need to copy or re-partition parameters at runtime, and simplifies the interface between scheduling and communication.

\subsection{2D Parallelism: Combining PertP and DDP} \label{sec:2d-parallel}

While Perturbation Parallelism (PertP) improves training throughput by distributing dual forward passes across GPUs, and Distributed Data Parallelism (DDP) enables model replication with efficient scalar synchronization, each strategy alone has limitations. PertP is constrained to two-way parallelism and does not scale beyond two GPUs without further extension, while DDP suffers from underutilization of per-GPU resources when each device must redundantly execute both perturbed computations.

\begin{wrapfigure}{r}{0.7\textwidth}
\vspace{-0.2in}
  \begin{center}
    \includegraphics[width=0.7\textwidth]{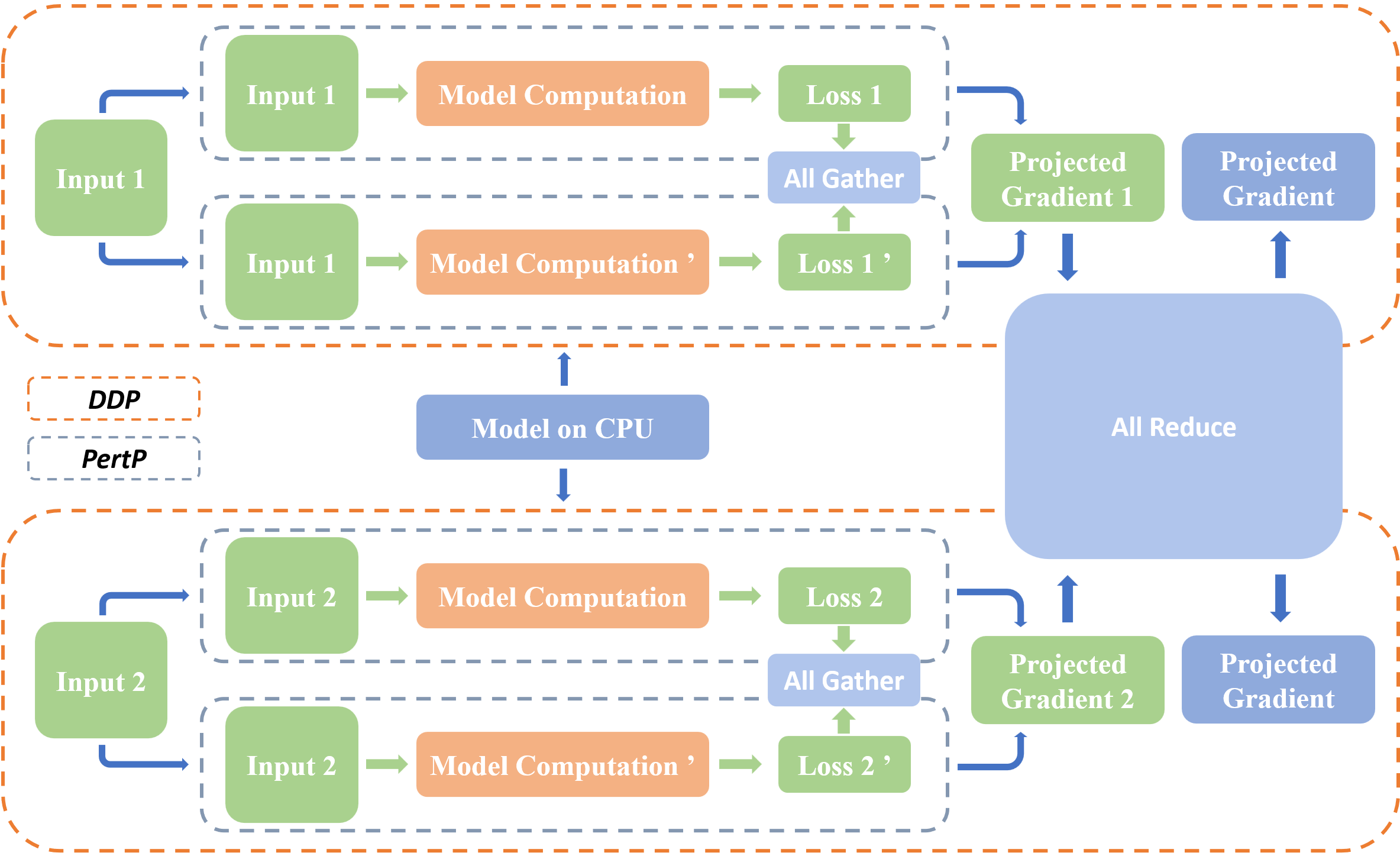}
  \end{center}
  \caption{2D Parallelism: inner PertP and outer DDP.}
  \label{fig:2d-parallel}
\vspace{-0.1in}
\end{wrapfigure}

To scale zeroth-order optimization efficiently across more number of GPUs, we propose to combine PertP and DDP into a unified \textbf{2D parallelism framework}, where each GPU performs only a single forward pass (with either $+\epsilon z$ or $-\epsilon z$) on a unique data shard. This fusion allows simultaneous parallelism along two orthogonal dimensions: \textit{perturbation direction} and \textit{data batches}.

We organize the total set of $n$ GPUs as a 2D mesh of size $n = n_b \times n_p$, where $n_b$ is the number of DDP batch-parallel groups and $n_p = 2$ is the perturbation dimension. For each mini-batch, we perform the following steps:

\textbf{(1) Group Assignment.} Each GPU is assigned a unique tuple $(b_i, p_j)$, where $b_i$ denotes the data-parallel group and $p_j \in \{+\epsilon, -\epsilon\}$ indicates the perturbation direction. \\
\textbf{(2) Forward Execution.} Each GPU computes a forward pass for its assigned perturbation on its local batch partition. No GPU computes both directions. \\
\textbf{(3) Intra-Group Communication.} Within each DDP group (fixed $b_i$), the pair of GPUs corresponding to $+\epsilon$ and $-\epsilon$ exchange their scalar losses $L^+$ and $L^-$ to compute a shared projected gradient:
    $g_i = \frac{L^+ - L^-}{2\epsilon}$. \\
\textbf{(4) Global Gradient Synchronization.} The projected gradients $\{g_i\}$ from all $n_b$ groups are synchronized across all devices using \texttt{all\_reduce}, yielding the final global gradient $g$. \\
\textbf{(5) Parameter Update.} Each GPU independently updates its parameters using $g$ and the shared perturbation vector $z$, ensuring consistency across the entire model replica.

This design uses PertP as the \textit{inner parallelism} within each DDP group and DDP as the \textit{outer parallelism} across batches. This structure minimizes communication overhead, aligns with standard DDP usage, and scales efficiently. In contrast, reversing the order—using DDP as the inner parallelism—introduces unnecessary cross-group synchronization, complicates RNG coordination, and deviates from typical DDP frameworks, without offering performance gains. Thus, the PertP-inner structure is both simpler and more effective.

\section{Experiments}

We evaluate our framework using the Open Pre-trained Transformer (OPT) \citep{zhang2022opt} model family, including sizes ranging from 1.3B to 175B, to assess scalability across model scales. All experiments are conducted on the SST-2 \citep{socher2013recursive} dataset using the MeZO \citep{malladi2023fine} and ZO2 \citep{wang2025zo2scalablezerothorderfinetuning} methods as baselines. Our implementation builds upon ZO2 and extends it to multi-GPU settings.

Experiments are performed on a server with 4×NVIDIA H20 GPUs (96GB each) and an AMD Milan CPU, using PyTorch 2.3.0, CUDA 12.7, and Python 3.12. We use consistent hyperparameters across all runs: learning rate $1\mathrm{e}{-7}$, batch size (BS) 4, sequence length 2048, and 100 training steps. GPU memory usage per GPU and training throughput (tokens/second) are reported for each configuration. Our implementation leverages CUDA streams for overlapping upload, compute, and offload. 

\subsection{Main Experiment}

The goal of our main experiment is to evaluate the effectiveness of our distributed strategies under a fixed batch size setting. Specifically, we compare DistZO2 variants—including Perturbation Parallelism (PertP) and Distributed Data Parallelism (DDP) using 2 GPUs, and the unified PertP(inner)+DDP(outer) configuration using 4 GPUs—against single-GPU baselines (MeZO and ZO2).

\begin{table*}[h]
\centering
\caption{\textbf{Main results of ZO2 performance for various model configurations and both FP32 and FP16 modes.} Instances of `-' in the table indicate scenarios where the corresponding method failed to execute due to memory constraints.}
\resizebox{\textwidth}{!}{
\begin{tabular}{cccccc|ccccc}
\toprule
\multirow{2}{*}{Model} & \multicolumn{5}{c|}{Memory Usage (MB) / GPU $\downarrow$} & \multicolumn{5}{c}{Throughput (tokens/sec) $\uparrow$} \\ 
\cline{2-11}
& \makecell{MeZO} & \makecell{ZO2} & \makecell{ZO2\\+\\PertP} & \makecell{ZO2\\+\\DDP} & \makecell{ZO2\\+PertP\\+DDP}
& MeZO & ZO2 & \makecell{ZO2\\+\\PertP} & \makecell{ZO2\\+\\DDP} & \makecell{ZO2\\+PertP\\+DDP} \\
\midrule 
OPT-1.3B & 5877 & 4979 & 5627 & 4451 & 6275 & 18720 & 18625 & 34225 & 31585 & 60091 \\

OPT-2.7B & 8617 & 5419 & 6021 & 4845 & 6669 & 9582 & 9582 & 17720 & 16106 & 30918 \\

OPT-6.7B & 16791 & 6941 & 7393 & 5823 & 8041 & 4133 & 4133 & 7653 & 7062 & 14417 \\ 

OPT-13B & 28999 & 8103 & 8455 & 7081 & 9201 & 2197 & 2190 & 4091 & 3730 & 7057 \\ 

OPT-30B & 62435 & 10723 & 10883 & 9901 & 11825 & 976 & 976 & 1820 & 1661 & 3221 \\ 

OPT-66B & - & 13263 & 13735 & 12163 & 14775 & - & 446 & 885 & 764 & 1370 \\ 

OPT-175B & - & 18713 & 18829 & 19881 & 18713 & - & 166 & 331 & 290 & 508\\ 
\bottomrule
\end{tabular}
}
\label{tab:exp-mainresult-performance}
\end{table*}

Table~\ref{tab:exp-mainresult-performance} summarizes results across OPT model scales from 1.3B to 175B parameters. We make the following observations: \textbf{(1) Memory Efficiency.} All DistZO2 variants preserve the memory efficiency of ZO2. Notably, PertP and DDP maintain similar or lower per-GPU memory usage compared to ZO2, despite involving additional computation or communication. \textbf{(2) Throughput Gains from PertP and DDP.} Introducing Perturbation Parallelism (ZO2+PertP) achieves up to $2\times$ speedup across all models by parallelizing the dual forward passes. Distributed Data Parallelism (ZO2+DDP) provides consistent throughput improvements by scaling across mini-batches. ZO2+PertP consistently achieves slightly higher throughput than ZO2+DDP.
\textbf{(3) Best Performance with Unified 2D Parallelism.} The combination of PertP and DDP (ZO2+PertP+DDP) delivers the highest throughput across all configurations. On OPT-175B, our unified parallelism achieves a $3\times$ speedup over ZO2.
\textbf{(4) Scaling to Extremely Large Models.} MeZO fails to run models beyond OPT-30B due to memory limitations, while DistZO2 successfully fine-tunes up to OPT-175B using only 18–20GB GPU memory per device.

\subsection{Ablation Study}

We conduct a series of ablation studies to isolate and quantify the contributions of each design component in DistZO2. Specifically, we analyze the effect of Perturbation Parallelism (PertP) (Section~\ref{sec:pertp}), Distributed Data Parallelism (DDP) (Section~\ref{sec:ddp}), our hardware-aware communication optimization (Section~\ref{sec:comm}), and their combined form in 2D parallelism (Section~\ref{sec:2d-parallel}). All experiments use a fixed BS (2, 4, 6, 8) and are conducted across multiple model scales to ensure generality. Although our distributed strategies are primarily designed to accelerate ZO2, they are orthogonal to the memory management style of the optimizer. Since MeZO and ZO2 share the same forward-pass structure and scalar-gradient computation, our methods can be directly applied to MeZO. This also allows for more controlled comparisons when isolating the effects of parallelism alone.

\begin{table}[h]
\centering
\caption{\textbf{Throughput (Tokens per Second) Analysis Across Different Parallelism Methods.}}
\resizebox{0.95\textwidth}{!}{
\begin{tabular}{cc|cc|cc|cc}
\toprule
Model & BS & \shortstack{MeZO\\+PertP}  & \shortstack{ZO2\\+PertP}  & \shortstack{MeZO\\+DDP}  & \shortstack{ZO2\\+DDP}  & \shortstack{ZO2\\+DDP+PertP} & \shortstack{ZO2\\+PertP+DDP}  \\
\midrule
OPT-1.3B &  & 28196 & 28196 & 24160 & 24160 & 32364 & 33815 \\
OPT-2.7B &  & 14512 & 14512 & 12426 & 12394 & 20837 & 20818 \\
OPT-6.7B & 2 & 6342 & 6342 & 5622 & 5391 & 6624 & 9227 \\
OPT-13B &  & 3369 & 3369 & 2982 & 2925 & 3747 & 4883 \\
OPT-30B &  & 1500 & 1500 & 1329 & 1329 & 2193 & 2238 \\
\midrule
OPT-1.3B &  & 34416 & 34225 & 31772 & 31585 & 54343 & 60091 \\
OPT-2.7B &  & 17746 & 17720 & 16148 & 16106 & 28998 & 30918 \\
OPT-6.7B & 4 & 7653 & 7653 & 7062 & 7062 & 12669 & 14417 \\
OPT-13B &  & 4091 & 4091 & 3730 & 3730 & 6676 & 7057 \\
OPT-30B &  & 1820 & 1820 & 1661 & 1661 & 3079 & 3221 \\
\midrule
OPT-1.3B &  & 37643 & 36965 & 35375 & 35375 & 64031 & 67212 \\
OPT-2.7B &  & 19035 & 18904 & 17888 & 17823 & 32874 & 33943 \\
OPT-6.7B & 6 & 8305 & 8184 & 7835 & 7653 & 14406 & 15346 \\
OPT-13B &  & 4407 & 4398 & 4104 & 4104 & 7592 & 7962 \\
OPT-30B &  & 1936 & 1936 & 1812 & 1812 & 3343 & 3350 \\
\midrule
OPT-1.3B &  & 38869 & 37777 & 36892 & 36744 & 68623 & 72164 \\
OPT-2.7B &  & 19907 & 19573 & 18930 & 18830 & 35544 & 36414 \\
OPT-6.7B & 8 & 8582 & 8522 & 8122 & 8122 & 15251 & 15944 \\
OPT-13B &  & 4584 & 4557 & 4325 & 4325 & 8142 & 8487 \\
OPT-30B &  & 2015 & 2001 & 1924 & 1924 & 3618 & 3802 \\
\bottomrule
\end{tabular}
}
\label{tab:ablation-throughput-analysis}
\end{table}

\begin{table}[h]
\centering
\caption{\textbf{Throughput Analysis (on 4 GPUs): Communication Optimization (Billion Parameters per Second).}}
\resizebox{0.9\textwidth}{!}{
\begin{tabular}{c|cc|cc}
\toprule
Model & Upload & Upload+CommOpt & Offload  & Offload+CommOpt \\ 
\midrule
OPT-1.3B  & 4.73  & 19.24 & 1.79  & 8.32 \\
OPT-2.7B  & 4.67  & 18.35 & 1.42  & 6.76 \\
OPT-6.7B  & 4.33  & 12.06 & 2.00  & 7.92 \\ 
OPT-13B   & 4.45  & 17.75 & 1.81  & 8.29 \\ 
OPT-30B   & 3.69  & 17.71 & 1.98  & 8.80 \\ 
OPT-66B   & 4.08  & 14.20 & 1.99  & 7.06 \\ 
OPT-175B  & 3.82  & 17.66 & 2.12  & 8.77 \\ 
\bottomrule
\end{tabular}
}
\label{tab:ablation-comm}
\end{table}

\textbf{Impact of Perturbation Parallelism (PertP) or Distributed Data Parallelism (DDP).}
Table~\ref{tab:ablation-throughput-analysis} Columns 3-6 show that applying PertP or DDP allows ZO2 to maintain throughput comparable to MeZO across all model scales. This indicates that our distributed execution strategies effectively offset the inherent performance cost of ZO2's CPU offloading and block-wise scheduling across all BSs. In other words, even though ZO2 trades computation speed for memory efficiency, PertP and DDP restore its throughput to match that of MeZO without compromising scalability or memory savings.

\textbf{Combined 2D Parallelism Efficiency.}
We combine PertP and DDP into a unified 2D parallelism framework, where each GPU performs a single forward pass on a disjoint data shard. Table~\ref{tab:ablation-throughput-analysis}, Columns 7–8, compare two design variants: (1) \textit{DDP(inner) + PertP(outer)} and (2) \textit{PertP(inner) + DDP(outer)}. 
Across all model sizes on 4 GPUs, the \textit{PertP(inner) + DDP(outer)} strategy consistently achieves higher throughput across all BSs. This result empirically confirms our design choice and supports the theoretical justification presented in Section~\ref{sec:2d-parallel}, where we argue that minimizing cross-branch communication and maintaining standard DDP compatibility makes PertP(inner) the preferable configuration.

\textbf{Effect of Communication Optimization.}
To reduce PCIe bottlenecks, we introduce a communication strategy that slices parameter blocks and redistributes them via NVLink. Table~\ref{tab:ablation-comm} reports upload and offload throughput (in billion parameters/sec) before and after applying our optimization. 
We isolate communication bandwidth rather than overall training throughput, since computation may still dominate end-to-end runtime. Nevertheless, our optimization achieves up to $4.6\times$ improvement in both upload and offload speeds, which is essential for enabling effective overlap with forward computation in multi-GPU execution.

\section{Conclusion}

We present \textbf{DistZO2}, a high-throughput and memory-efficient framework for zeroth-order fine-tuning of large language models. Building upon the memory-saving design of ZO2, DistZO2 introduces three distributed strategies—Perturbation Parallelism, Distributed Data Parallelism, and their unified 2D parallelism—that significantly improve throughput without increasing GPU memory usage.
Through extensive experiments on models up to OPT-175B, we show that DistZO2 achieves up to $3\times$ speedup over ZO2 and restores throughput to the level of MeZO, while maintaining the ability to run on GPUs with less than 20GB memory. Our hardware-aware communication optimization further enhances multi-GPU efficiency, enabling scalable and practical fine-tuning for multi-hundred-billion-parameter models.

\bibliography{main}
\bibliographystyle{colm2025_conference}

\newpage
\appendix
\section{More Figures, Algorithms, and Tables}

\begin{algorithm}
\caption{MeZO \citep{malladi2023fine}}
\label{alg:mezo}
\begin{algorithmic}[1]
\Require Model parameters $\theta \in \mathbb{R}^d$, loss function $L : \mathbb{R}^d \rightarrow \mathbb{R}$, training iterations $T$, perturbation step size $\epsilon$, data batch $B$, learning rate $\eta_t$
\For{$j = 1, \dots, T$}
    \State Set random seed $s$ and sample batch $B \subset D$
    \State $\theta \leftarrow \text{PerturbParameters}(\theta, \epsilon)$
    \State $\ell_+ \leftarrow L(\theta; B)$
    \State Reset RNG with seed $s$
    \State $\theta \leftarrow \text{PerturbParameters}(\theta, -2\epsilon)$
    \State $\ell_- \leftarrow L(\theta; B)$
    \State Reset RNG with seed $s$
    \State $\theta \leftarrow \text{PerturbParameters}(\theta, \epsilon)$
    \State $g \leftarrow (\ell_+ - \ell_-) / (2\epsilon)$
    \State Reset RNG with seed $s$
    \State $\theta \leftarrow \text{UpdateParameters}(\theta, g)$
\EndFor
\Statex
\Function{UpdateParameters}{$\theta, g$}
    \For{each $\theta_i \in \theta$, where $\theta_i \in \mathbb{R}^{d_i}$}
        \State $z_i \sim \mathcal{N}(0,1) \in \mathbb{R}^{d_i}$
        \State $\theta_i \leftarrow \theta_i - \eta_t \cdot g \cdot z_i$
    \EndFor
    \State \Return $\theta$
\EndFunction
\Statex
\Function{PerturbParameters}{$\theta, \epsilon$}
    \For{each $\theta_i \in \theta$, where $\theta_i \in \mathbb{R}^{d_i}$}
        \State $z_i \sim \mathcal{N}(0,1) \in \mathbb{R}^{d_i}$
        \State $\theta_i \leftarrow \theta_i + \epsilon z_i$
    \EndFor
    \State \Return $\theta$
\EndFunction
\end{algorithmic}
\end{algorithm}

\begin{figure}[htbp]
    \centering
    \includegraphics[width=\linewidth]{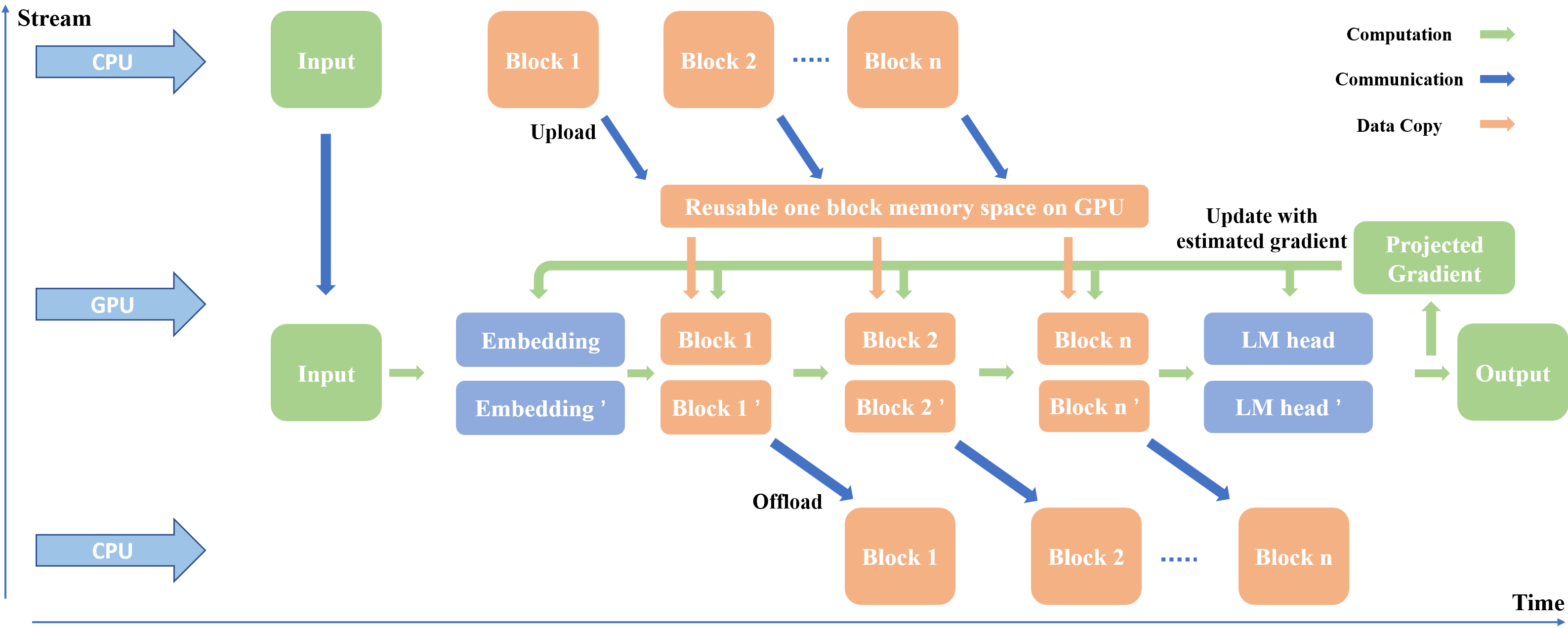}
    \caption{ZO2 \citep{wang2025zo2scalablezerothorderfinetuning} workflow.}
    \label{fig:zo2}
\end{figure}

\begin{algorithm}
\caption{ZO2 \citep{wang2025zo2scalablezerothorderfinetuning} Computation with RNG State Manager}
\label{alg:zo2}
\begin{algorithmic}[1]
\Require{Transformer blocks $\{W_i\}_{i=1}^N$ with number of transformer blocks $N$, embedding parameters $Embedding$, and LM head $LM head$, module parameter $\theta$, module forward function $forward$, loss function $L$, training iterations $T$, perturbation step size $\epsilon$, data batch $B$, learning rate $\eta_t$, random seed $s$, random state buffer $rsb$, last iteration's random state $lrs$.}
\State Initialize $g=0$.
\For{$j = 1, \dots, T$}
    \State Set random seed $s$ and sample batch $B \subset D$
    \State Get random state $rs$ = GetRngState($s$), and push $rs$ into $rsb$.
    \If{$j$ > 1}
        \State Update last iteration's random state $lrs$ = PopLeft($rsb$)
    \Else
        \State $lrs$ = $None$
    \EndIf
    \State $out_+, out_-, rs, lrs$ = DualForward(
        \Statex $Embedding, \epsilon, s, rs, lrs, g, B, B$)
    \For{$i=1$ to $N$}
        \State $out_+, out_-, rs, lrs$ = DualForward(
        \Statex $W_i, \epsilon, s, rs, lrs, g, out_+, out_-$)
    \EndFor
    \State $out_+, out_-, rs, lrs$ = DualForward(
    \Statex $LM head, \epsilon, s, rs, lrs, g, out_+, out_-$)
    \State $\ell_+ = L(out_+)$, $\ell_- = L(out_-)$
    \State $g \leftarrow (\ell_+ - \ell_-) / (2\epsilon)$
\EndFor
\Statex
\Function{DualForward}{$\theta, \epsilon, s, rs, lrs, g, input_+, input_-$}
    \If{$g$ != 0}
        \State SetRngState($s, lrs$)
        \State $\theta, lrs \leftarrow$ UpdateParameters($\theta, g$)
    \EndIf
    \State SetRngState($s, rs$), $\theta \leftarrow \text{PerturbParameters}(\theta, \epsilon)$
    \State $out_+ \leftarrow forward(\theta; input_+)$
    \State SetRngState($s, rs$), $\theta \leftarrow \text{PerturbParameters}(\theta, -2\epsilon)$
    \State $out_- \leftarrow forward(\theta; input_-)$
    \State SetRngState($s, rs$), $\theta \leftarrow \text{PerturbParameters}(\theta, \epsilon)$
    \State $rs$ = GetRngState($s$)
    \State \Return $out_+, out_-, rs, lrs$
\EndFunction
\Statex
\Function{SetRngState}{$s, rs$}
    \State Store $rs$ associated with seed $s$. 
\EndFunction
\Statex
\Function{GetRngState}{$s$}
    \State \Return $rs$ associated with seed $s$ from the storage.
\EndFunction
\Statex
\Function{UpdateParameters}{$\theta, g$}
    \State Same as Algorithm \ref{alg:mezo}'s UpdateParameters
\EndFunction
\Statex
\Function{PerturbParameters}{$\theta, \epsilon$}
    \State Same as Algorithm \ref{alg:mezo}'s PerturbParameters
\EndFunction
\end{algorithmic}
\end{algorithm}

\begin{algorithm}[htbp]
\caption{ZO2 \citep{wang2025zo2scalablezerothorderfinetuning} Dynamic Scheduler}
\label{alg:zo2-scheduler}
\begin{algorithmic}[1]
\Require{Transformer blocks $\{W_i\}_{i=1}^N$ with number of transformer blocks $N$, embedding parameters $Embedding$, and LM head $LM head$}.
\State Initialize a dynamic scheduler $S\{ \cdot \}$ to control dual forward computation $C( \cdot )$, uploading $U( \cdot )$, and offloading $O( \cdot )$ operations.
\State Asynchronously launch $S\{C(Embedding), U(W_1)\}$.
\For{$i=1$ to $N-1$}
    \State Synchronously wait until $U(W_i)$ finished.
    \If{$i=1$}
        \State Asynchronously launch $S\{C(W_{i}), U(W_{i+1})\}$.
    \Else
        \State Synchronously wait until $C(W_{i-1})$ finished.
        \State Asynchronously launch $S\{O(W_{i-1}), C(W_{i}), U(W_{i+1})\}$.
    \EndIf
\EndFor
\State Synchronously wait until $C(W_{N-1})$ and $U(W_N)$ finished.
\State Asynchronously launch $S\{O(W_{N-1}), C(W_N)\}$.
\State Synchronously wait until $C(W_N)$ finished.
\State Asynchronously launch $S\{O(W_N), C(LM head)\}$.
\end{algorithmic}
\end{algorithm}

\section{Full Related Work} \label{sec:related-work}

Zeroth-order (ZO) optimization has emerged as a memory-efficient alternative for fine-tuning large language models (LLMs), as it avoids backward passes by estimating gradients using forward-only computations. MeZO~\citep{malladi2023fine} and ZO2~\citep{wang2025zo2scalablezerothorderfinetuning} demonstrate that ZO methods can significantly reduce memory usage and enable tuning of multi-hundred-billion-parameter models on a single GPU. However, these methods suffer from low throughput and poor scalability. 

\subsection{Zeroth-Order Optimization}

Zeroth-order (ZO) optimization provides a gradient-free alternative to first-order methods by estimating gradients solely from function evaluations. Foundational approaches such as random perturbation-based gradient estimation~\citep{flaxman2004online} and two-point finite differences~\citep{nesterov2017random} have established theoretical guarantees under smoothness assumptions. ZO methods have found applications in black-box adversarial attacks~\citep{chen2017zoo}, reinforcement learning~\citep{salimans2017evolution}, and model-agnostic optimization.

Despite scalability concerns due to dimension-dependent query complexity, techniques such as block-coordinate estimation~\citep{cai2021zeroth} and sparsity-aware gradient estimation~\citep{cai2022zeroth} have been proposed to improve efficiency in high-dimensional settings.

\subsection{Zeroth-Order Optimization for LLM Fine-Tuning}

As large language models (LLMs) grow in scale, memory-efficient fine-tuning becomes increasingly critical. MeZO~\citep{malladi2023fine} introduced a backpropagation-free fine-tuning framework using ZO-SGD, enabling forward-only updates with drastically reduced memory footprint. This approach eliminates the need to store activation states and gradient tensors, allowing large-scale models to be fine-tuned on commodity hardware.

Subsequent work~\citep{zhang2024revisitingzerothorderoptimizationmemoryefficient} expanded the study of ZO methods by benchmarking a wide range of optimizers, including ZO-Adam, momentum-based ZO-SGD, and block-wise strategies. These improvements yield better stability and convergence, although most assume the full model fits in GPU memory.

More recently, ZO2~\citep{wang2025zo2scalablezerothorderfinetuning} demonstrated that ZO optimization is particularly well-suited for CPU offloading due to its forward-only execution pattern. ZO2 introduces system-level innovations such as an RNG state manager, a lightweight CUDA stream-based scheduler, and AMP-aware low-bit communication. These enable efficient fine-tuning of models up to 175B parameters using just 18GB of GPU memory, without sacrificing accuracy or throughput.

Together, these works show that ZO-based fine-tuning, once considered too inefficient for large-scale models, is becoming a practical solution for memory-constrained LLM training.

\subsection{Distributed Computing}

Meanwhile, distributed training strategies such as Distributed Data Parallelism (DDP)~\citep{li2020pytorchdistributedexperiencesaccelerating} is widely used in first-order training~\citep{ruder2017overviewgradientdescentoptimization, loshchilov2017decoupled}, but require adaptation for ZO due to its scalar-gradient nature and dual forward passes. In particular, DDP must synchronize scalar gradients and coordinate consistent perturbation vectors across devices. These adaptations are non-trivial and introduce scheduling and communication challenges.

To address these, DistZO2 first introduces Perturbation Parallelism (PertP), and then adapts DDP to support ZO training. Furthermore, it proposes a unified 2D parallelism framework that combines PertP with DDP, enabling parallelism along both perturbation and data dimensions. It further proposes a hardware-aware communication strategy that slices parameter blocks and redistributes them via fast interconnects like NVLink to alleviate PCIe bottlenecks. These innovations allow DistZO2 to significantly improve throughput while preserving ZO2’s memory efficiency. Our approach is also compatible with recent forward acceleration techniques such as FlashAttention~\citep{dao2023flashattention2fasterattentionbetter} and \texttt{torch.compile}~\citep{ansel2024pytorch}, which further emphasize the need for efficient communication in forward-only training.

\section{More Experiments}

\subsection{Additional Scaling Results with GPU Count}

\begin{table}[htbp]
\centering
\caption{\textbf{Throughput (Tokens/Second) Comparison Across Models and GPU Counts.} Batch size is fixed at 4, using fp16 precision.}
\resizebox{0.6\textwidth}{!}{
\begin{tabular}{lc|cc}
\toprule
Model & Num GPUs & MeZO+DDP & ZO2+DDP\\
\midrule
OPT-1.3B &   & 18509 & 18446 \\ 
OPT-2.7B &    & 9485 & 9566 \\
OPT-6.7B & 1   & 4062 & 4060 \\
OPT-13B  &    & 2177 & 2179 \\
OPT-30B  &    & 968 & 961 \\
\midrule
OPT-1.3B &   & 31772 & 31585 \\
OPT-2.7B &    & 16148 & 16106 \\
OPT-6.7B & 2   & 7062 & 7062 \\
OPT-13B  &    & 3730 & 3730 \\
OPT-30B  &    & 1661 & 1661 \\
\midrule
OPT-1.3B &   & 48041 & 45684 \\
OPT-2.7B &    & 24802 & 24640 \\
OPT-6.7B & 4   & 11234 & 11284 \\
OPT-13B  &    & 5966 & 5955 \\
OPT-30B  &    & 2658 & 2637 \\
\bottomrule
\end{tabular}
}
\label{tab:throughput-gpu-ddp} 
\end{table}

To further validate the scalability of DistZO2, we compare MeZO+DDP and ZO2+DDP across 1, 2 and 4 GPUs while fixing the batch size to 4 and using fp16 precision. As shown in Table~\ref{tab:throughput-gpu-ddp}, both methods scale consistently with GPU count, achieving nearly linear throughput improvements from 1 to 4 GPUs. Importantly, the performance gap between MeZO and ZO2 remains negligible throughout, confirming that ZO2’s offloading mechanism does not introduce any significant parallelization overhead when using DDP.

\subsection{Sequence Length Sensitivity Across Parallelism Strategies}

\begin{table}[htbp]
\centering
\caption{\textbf{Throughput (Tokens per Second) Analysis Across Different Parallelism Methods.}}
\resizebox{0.9\textwidth}{!}{
\begin{tabular}{cc|ccccc}
\toprule
Model & \shortstack{Sequence\\Length} & \shortstack{ZO2\\+PerP}  & \shortstack{MeZO\\+DDP}  & \shortstack{ZO2\\+DDP}  & \shortstack{ZO2\\+DDP+PertP} & \shortstack{ZO2\\+PertP+DDP}  \\
\midrule
OPT-1.3B & & 29701 & 24845 & 24194  & 33796 & 32608 \\
OPT-2.7B &  & 15110 & 12850 & 12399 & 11070 & 10231\\
OPT-6.7B & 1024  & 7250 & 5738 & 5711 & 5151 & 5351\\
OPT-13B &  & 2118 & 3034 & 2996 & 2807 & 2658 \\
OPT-30B &  & 1703 & 1344 & 1344 & 1119 & 1239\\
\midrule

OPT-1.3B &   & 34225 & 31772 & 31585 & 54343 & 60091 \\
OPT-2.7B &   & 17720 & 16148 & 16106 & 28998 & 30918 \\
OPT-6.7B & 2048  & 7653 & 7062 & 7062 & 12669 & 14417 \\
OPT-13B &   & 4091 & 3730 & 3770 & 6676 & 7057 \\
OPT-30B &   & 1820 & 1661 & 1661 & 3079 & 3221 \\
\midrule

OPT-1.3B & & 35008 & 33762 & 33465 & 63377 & 44876  \\
OPT-2.7B &  & 17983 & 17326 & 17234 & 22558 & 24713\\
OPT-6.7B & 4096 & 8077& 7715 & 7763 & 11033 & 10941\\
OPT-13B &  & 4327 & 4147 & 4135 & 5917 & 5943 \\
OPT-30B &  & 1920 & 1863 & 1860 & 2657 & 2625\\
\midrule

OPT-1.3B & & 31948 & 31269 & 31236  & 61766 & 51007 \\
OPT-2.7B &  & 16231 & 15955 & 15942 & 26245 & 26741 \\
OPT-6.7B & 8192 & 7782 & 7609 & 7625 & 12581 & 12668 \\
OPT-13B &  & 4218 & 4145 & 4139 & 6867 & 6886 \\
OPT-30B &   & 1780 & 1883 & 1874 & 3116 & 3100\\
\bottomrule
\end{tabular}
}
\label{tab:throughput-seq-length}
\end{table}

We also evaluate how different parallelism strategies perform under varying sequence lengths, ranging from 1024 to 8192. As shown in Table~\ref{tab:throughput-seq-length}, all methods experience reduced throughput with longer sequences, as expected. However, DistZO2 maintains strong relative performance across all sequence lengths, with the unified 2D strategy (ZO2+PertP+DDP) consistently achieving the highest throughput.

\end{document}